**TITLE PAGE**

# Enhanced Classroom Dialogue Sequences Analysis with a Hybrid AI Agent: Merging Expert Rule-Base with Large Language Models

Yun Long[1], Yu Zhang[1*]

[1] Institute of Education, Tsinghua University, Beijing, China, 100084

[*] Corresponding author: Yu Zhang, zhangyu2011@mail.tsinghua.com, +86 10 62785686 (Tel)

## Acknowledgements

This work was supported by the Beijing Educational Science Foundation of the Fourteenth 5-year Planning (BGEA23019) and the National Natural Science Foundation of China (62177030).

## Declaration of Interest statement

The authors report there are no competing interests to declare.

## Corresponding author address

417 Wennan Building, Tsinghua University, Beijing, China, 100084

## Corresponding author email address

zhangyu2011@tsinghua.edu.cn

**Abstract**

Classroom dialogue plays a crucial role in fostering student engagement and deeper learning. However, analysing dialogue sequences has traditionally relied on either theoretical frameworks or empirical descriptions of practice, with limited integration between the two. This study addresses this gap by developing a comprehensive rule base of dialogue sequences and an Artificial Intelligence (AI) agent that combines expert-informed rule-based systems with a large language model (LLM). The agent applies expert knowledge while adapting to the complexities of natural language, enabling accurate and flexible categorisation of classroom dialogue sequences. By synthesising findings from over 30 studies, we established a comprehensive framework for dialogue analysis. The agent was validated against human expert coding, achieving high levels of precision and reliability. The results demonstrate that the agent provides theory-grounded and adaptive functions, tremendously enhancing the efficiency and scalability of classroom dialogue analysis, offering significant potential in improving classroom teaching practices and supporting teacher professional development.

## 1. Introduction

Classroom dialogue plays a crucial role in fostering student engagement, knowledge construction, and critical thinking (Alexander, 2008; Mercer & Dawes, 2014). High-quality dialogue enables students to articulate their thoughts, reason through problems, and collaboratively build on each other's ideas—key processes for deep learning (Howe & Abedin, 2013). Analysing classroom dialogue is therefore essential for assessing dialogue quality and providing educators with insights to refine their teaching practices. Through structured analysis, researchers and practitioners can evaluate how well classroom interactions align with theoretical models of effective dialogue, ultimately bridging the gap between theory and practice to improve learning outcomes.

Despite its importance, the current landscape of classroom dialogue analysis remains fragmented. Most studies fall into two primary domains: theoretical analysis and empirical investigation. Theoretical analyses focus on identifying ideal dialogue sequences that align with pedagogical goals, such as fostering exploratory talk or critical inquiry (Mercer & Wegerif, 1999; Howe et al., 2019). These studies provide normative models that define what classroom dialogue *should* look like. Conversely, empirical research emphasises describing *actual* dialogue patterns observed in real classrooms, often revealing significant deviations from theoretical ideals (Nystrand et al., 2003). However, these data-driven studies tend to prioritise empirical accuracy over theoretical alignment, leaving educators with limited actionable strategies for bridging the gap between ideal and real-world dialogue. Consequently, translating theoretical insights into practice remains a persistent challenge.

The advent of AI offers promising solutions to these challenges. AI approaches can broadly be categorised into three paradigms: symbolic AI, narrow connectionist AI, and general connectionist AI (McCarthy et al., 1955; LeCun et al., 2015; Zhang et al., 2020), each offering distinct advantages for classroom dialogue analysis. Symbolic AI, characterised by rule-based systems, excels in applying explicit expert knowledge to structured tasks, such as identifying specific dialogue patterns like Initiate-Response-Evaluate (IRE) sequences (Buchanan & Shortliffe, 1984; Mercer, 1995). Narrow connectionist AI, such as convolutional neural networks (CNNs) and recurrent neural networks (RNNs), has demonstrated effectiveness in identifying dialogue moves like elaboration or reasoning from large datasets (LeCun et al., 2015; Shaffer et al., 2016). However, these models require extensive data collection and manual annotation, which are both costly and time-consuming. Moreover, the trained models often lack generalisability, limiting their applicability to other dialogue analysis tasks without significant retraining, thereby constraining their scalability and versatility. Meanwhile, general connectionist AI, exemplified by LLMs such as GPT-4, demonstrates exceptional adaptability in capturing the complexity of open-ended classroom dialogues, performing tasks like summarisation and topic shift detection with minimal supervision (Brown et al., 2020; Long et al., 2024). However, while symbolic AI ensures theoretical alignment, and narrow AI has traditionally been used for large-scale data processing, general connectionist AI, particularly LLMs, also demonstrates considerable capability in processing vast datasets. LLMs bring unmatched flexibility, allowing them to handle diverse dialogue contexts with minimal supervision. However, they often lack theoretical grounding, which can limit their interpretability and alignment with pedagogical frameworks. This underscores the need for a hybrid approach that combines the theoretical rigor of symbolic methods with the adaptability and scalability of LLMs.

Each paradigm offers unique strengths and limitations. This study develops a hybrid system combining the theoretical rigour of symbolic AI with the adaptability of general connectionist AI, specifically LLMs. By aligning dialogue analysis with expert knowledge while accommodating the complexities of natural language, this approach ensures both theoretical precision and contextual adaptability. This hybrid model not only improves the robustness of dialogue analysis by ensuring theoretical alignment and adaptability but also significantly enhances efficiency. By leveraging AI for Science, it transforms the scope, capacity, and speed of classroom dialogue analysis, enabling larger sample sizes and more generalisable conclusions. Moreover, it provides practical, theory-informed feedback to educators, bridging the gap between theoretical ideals and classroom realities while redefining research capabilities in this domain.

## 2. Literature Review

### 2.1 Classical approaches of classroom dialogue sequences analysis

The study of classroom dialogue sequences has long focused on two fundamental questions: (1) What are the ideal dialogue sequences that support different teaching goals, and (2) How do real-world classroom dialogue sequences compare to these ideal sequences? These two approaches have shaped much of the classical research on dialogue sequence analysis.

The first approach, theory-driven or deductive, examines ideal dialogue structures designed to achieve specific pedagogical goals. These studies often draw on well-established frameworks, such as Bloom's Taxonomy (Bloom, 1956) and the Cambridge Dialogue Analysis Scheme (CDAS) (Howe et al., 2019), to propose sequences that optimise learning. Examples include *exploratory talk*, which fosters critical engagement and deep learning (Mercer & Wegerif, 1999), and *cumulative talk*, where students build uncritically on each other's ideas to enhance mutual understanding (Mercer, 1995). Other notable types include *Socratic dialogue*, which uses systematic questioning to deepen understanding (Dillon, 1988), and *accountable talk*, emphasising rigorous thinking and coherence (Michaels et al., 2008). These frameworks identify ideal sequences, such as reasoning and elaboration chains, to promote higher-order thinking and collaborative knowledge construction, providing a theoretical blueprint for effective classroom interactions.

The second approach, data-driven or inductive, focuses on analysing real-world classroom dialogue sequences to determine how closely they align with the ideal patterns proposed by theory. Researchers in this domain collect and analyse dialogue data, seeking to understand whether actual classroom interactions meet the theoretical expectations for effective learning. Studies have revealed that real-world classroom dialogues often diverge from these ideal patterns, with teachers frequently controlling the flow of dialogue and students having limited opportunities for extended reasoning or questioning (Nystrand et al., 2003). This inductive analysis often highlights the gap between theory and practice in classroom discourse, underscoring the challenges of implementing idealised dialogue sequences in diverse educational contexts.

To perform such analyses, researchers frequently employ common sequence analysis methods such as Lag Sequential Analysis (LSA), Markov Chains, Process Mining, and Epistemic Network Analysis (ENA). Lag Sequential Analysis (LSA) has been widely used to identify temporal patterns in classroom interactions by examining the conditional probabilities of one event following another (Bakeman & Quera, 2011). This method is particularly valuable for understanding the structure of classroom dialogue by identifying recurring sequences of events, such as teacher questions followed by student responses. Markov Chains are often used to model the probabilistic transitions between different states in a dialogue sequence. By mapping the likelihood of moving from one dialogue act to another, Markov Chains provide insight into the underlying structure of classroom interactions and can be used to test how closely real dialogue follows ideal sequences (Rabiner, 1989). Process Mining has been applied to educational settings to trace the flow of activities and interactions over time. This method provides a visual representation of classroom dialogue processes and can identify deviations from expected patterns (Van der Aalst, 2016). Epistemic Network Analysis (ENA) is an emerging tool that maps the connections between different epistemic contributions in dialogue, revealing how students build on each other's ideas to construct knowledge. ENA is particularly useful for examining the deeper cognitive processes involved in collaborative learning and has been applied to classroom discourse to study the development of critical thinking and reasoning (Shaffer et al., 2016).

While these methods offer powerful tools for analysing dialogue sequences, their focus often differs in terms of theoretical grounding. Some studies explicitly compare real-world dialogue sequences to theoretical models, aiming to identify gaps or misalignments between ideal and actual classroom practices. However, such comparisons often remain descriptive, highlighting discrepancies without offering insights into how these gaps might be addressed in practice. On the other hand, many analyses are purely empirical, focusing on cataloguing the characteristics of real classroom dialogue without reference to theoretical frameworks. These studies, while valuable for understanding the practical dynamics of classroom interactions, lack a deeper theoretical lens to evaluate whether observed patterns align with pedagogical goals.

The primary limitation of both approaches lies in their inability to fully bridge the gap between theory and practice. Studies that emphasise theoretical comparison often fall short in providing actionable strategies for educators, while purely descriptive research offers limited guidance on how observed dialogue patterns could be improved to better align with theoretical ideals. This dual limitation underscores the need for methodologies that not only analyse real dialogue patterns but also provide theory-informed, actionable insights to support educators in refining their practice. By addressing these gaps, our research aims to develop a hybrid approach that integrates theoretical and empirical perspectives, offering a more comprehensive framework for classroom dialogue analysis.

### 2.2 AI-facilitated classroom dialogue analysis

Recent years have seen growing interest from leading scholars in the use of automated annotation technology in classroom research. Automated annotation refers to the use of machine learning algorithms to code transcribed classroom dialogues based on predefined codes, offering significant advantages in accuracy, speed, and scalability compared to traditional observation and statistical methods (Song, 2022). The process typically involves three stages: dataset creation, model training (a key part of machine learning), and automated coding. Among these, model training is pivotal, primarily relying on neural network algorithms (LeCun et al., 2015).

The common algorithms used for automated annotation include convolutional neural networks (CNNs), long short-term memory (LSTM) networks, and recurrent neural networks (RNNs), each excelling in different aspects of language processing. CNNs, for example, excel at capturing local semantic information, while LSTMs, as a more complex type of RNN, are designed to better handle long sequences of information. These technical advances not only improve the efficiency of dialogue analysis but also enable deeper insights into the underlying patterns of instructional practices. By analysing large-scale data, these algorithms can validate or refine existing pedagogical theories and uncover previously unrecognised dialogue features, contributing to the development of new theoretical frameworks.

Once the dialogue data is annotated, data mining techniques come into play to extract valuable patterns and relationships. Methods such as sequence pattern discovery, classification, clustering, and association rule mining are commonly applied (Wang et al., 2024). Sequence mining, in particular, holds unique advantages in identifying the temporal order of behaviours and speech, helping to uncover insights into how classroom discussions unfold over time (Santangelo, 2009). This type of analysis allows researchers to go beyond surface-level interactions such as question and response frequency, to investigate the deeper, cognitive processes that are influenced by dialogue sequences.

The first orientation is structure-based, focusing on identifying and categorising dialogue sequences using predefined theoretical frameworks. This approach often employs quantitative methods, such as Lag Sequential Analysis or Markov Chains, to capture the structural dynamics of dialogue patterns. Unlike previous studies that primarily use descriptive methods or rely on surface-level interaction features, these techniques delve deeper into the temporal and probabilistic relationships between dialogue moves. For instance, Lag Sequential Analysis examines the likelihood of specific dialogue moves following one another, while Markov Chains model the probabilistic transitions between states, offering a more systematic and predictive understanding of dialogue structure. This distinguishes them from more inductive, data-driven methods that lack a firm theoretical grounding. In contrast, the second orientation is function-based, where the focus is on the qualitative aspects of dialogue and its ability to foster thinking and reasoning. Recent efforts, like the CDAS (Howe et al., 2010) and the Teacher Scheme for Educational Dialogue Analysis (T-SEDA, Vrikki et al., 2018), emphasize dialogue features such as elaboration, reasoning, challenging and reflection. Other studies have applied NLP techniques to analyse group discussions, revealing patterns that enhance collaboration and knowledge construction (Wu et al., 2018; Song, 2022; Cukurova et al., 2018). Data mining and evidence-based methods have been employed to identify high-quality classroom behaviours, such as asking thought-provoking questions, but they often rely on post hoc analysis of data, missing the opportunity for real-time feedback (Wang et al., 2021).

The second orientation is function-based, where the focus is on the qualitative aspects of dialogue and its ability to foster thinking and reasoning. Recent efforts, like the CDAS (Howe et al., 2010) and the Teacher Scheme for Educational Dialogue Analysis (T-SEDA, Vrikki et al., 2018), emphasize dialogue features such as elaboration, reasoning, challenging and reflection. Other studies have applied NLP techniques to analyse group discussions, revealing patterns that enhance collaboration and knowledge construction (Wu et al., 2018; Song, 2022; Cukurova et al., 2018). Data mining and evidence-based methods have been employed to identify high-quality classroom behaviours, such as asking thought-provoking questions, but they often rely on post hoc analysis of data, missing the opportunity for real-time feedback (Wang et al., 2021).

Building upon these advancements, recent studies have explored the application of LLMs in classroom dialogue analysis, offering a promising solution to the challenges of manual annotation (authors, 2024). Traditional methods, while rich in depth and context, are often criticised for being knowledge- and labour-intensive, relying heavily on human expertise and qualitative coding, which are time-consuming and prone to subjectivity. A study by authors (2024) examined the potential of LLMs, particularly GPT-4, in analysing classroom dialogues to streamline and enhance the efficiency of educational research. Using datasets from middle school classrooms, including mathematics and Chinese lessons, the study compared expert manual annotations with outputs from a customised GPT-4 model, evaluating key factors such as time efficiency, inter-coder agreement, and reliability. Results showed that GPT-4 offered substantial time savings while maintaining a high degree of consistency with human coders. Although some discrepancies were observed in specific codes, overall inter-coder reliability was strong, indicating that LLMs could serve as reliable tools for reducing the burden of manual coding. These findings highlight the potential of LLMs to transform teaching evaluation and dialogue facilitation, providing not only a faster alternative to traditional methods but also a scalable solution for analysing large datasets (authors, 2024).

Despite these technological advancements, several limitations persist. One notable limitation is that LLMs have not been extensively applied to the recognition of dialogue sequences. Prior research has largely focused on extracting dialogue sequences using traditional machine learning and NLP techniques, typically through inductive logic—where patterns are inferred from the data to hypothesise possible sequences and their educational impacts. In some cases, these data-driven sequences are compared against theoretically ideal sequences, often drawn from frameworks such as exploratory talk or dialogic teaching (Mercer & Wegerif, 1999; Alexander, 2008). For example, studies may first use machine learning to identify common patterns in real-world classroom dialogues and then evaluate how closely these patterns align with normative models of effective dialogue. This comparison is typically done by calculating the frequency or probability of ideal sequences occurring in real settings and analysing deviations from these models. However, while this method provides some insights into the alignment between theory and practice, it still heavily relies on specific datasets and often lacks a comprehensive theoretical grounding in the automated process itself (Van Gog et al., 2019). As a result, the generalisability of findings is constrained, as these data-driven models may not adapt well to varied classroom contexts or fully leverage theoretical insights to guide automated analysis.

Therefore, this research aims to bridge the gap between the inductive, data-driven models of the past and a deductive framework grounded in expert knowledge. By validating dialogue sequences in real-world educational settings, we hope to offer more generalisable and theoretically sound conclusions about the nature of effective classroom dialogue, ultimately leading to more reliable and scalable improvements in teaching practices.

## 3. Research design and methods

To address the limitations of existing dialogue analysis approaches and bridge the gap between theoretical frameworks and empirical data, this study adopts a hybrid methodology. By combining expert-informed rule-based systems with the flexibility of LLMs, we aim to create an AI agent capable of accurately categorising and analysing classroom dialogue sequences in real time. This section outlines the research design, including the consolidation of existing dialogue types, the development of expert rules, and the implementation of the hybrid AI agent. The overall process is illustrated in Figure 1, which provides a visual representation of the study's methodological framework.

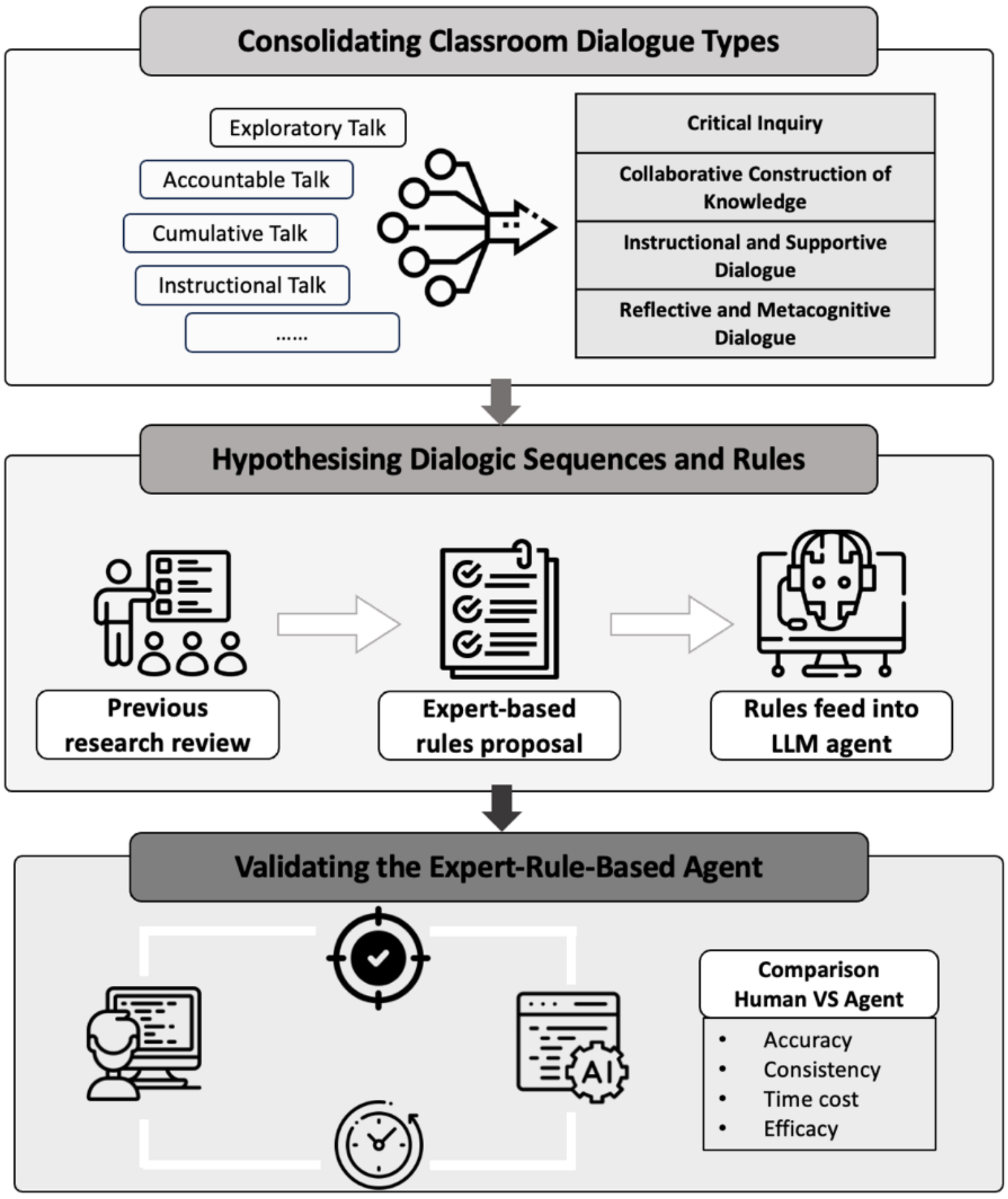

Figure 1. Research design diagram

### 3.1 Consolidation of Existing Classroom Dialogue Types

The initial step in our research design involves a thorough categorisation of existing dialogue types based on a comprehensive review of the literature on classroom dialogues (as shown in Figure 1). This review encompasses seminal works and recent studies that have identified various types of teacher-student and student-student interactions. From these studies, we synthesised the characteristics and patterns of effective classroom dialogues to form a consolidated categorisation. This categorisation serves as the foundational taxonomy necessary for defining new dialogue categories and their corresponding characteristics.

Then we define new dialogue categories and describe their unique characteristics. These definitions are based on the synthesised literature. We then employed the adapted coding scheme based on the Cambridge Dialogue Analysis Scheme (CDAS, Howe et al., 2019) to represent these newly defined categories. The adapted coding scheme includes categories such as Elaboration Invitation (ELI), Reasoning Invitation (REI), Elaboration (EL), Reasoning (RE), Coordination Invitation (CI), Simple

Coordination (SC), Reasoned Coordination (RC), Agreement (A), Querying (Q), Reference Back (RB), Reference to Wider Context (RW), Structural Silence (SU), Strategic Silence (SA) , Other Invitation (OI) and Other (O), details of this coding scheme is shown in Table 1.

**Table 1. The coding scheme used for analysing classroom dialogue***

| Codes | Brief definitions (and key words) |
|---|---|
| **Elaboration Invitation (ELI)** | Invites building on, elaboration, evaluation, clarification of own or another's contribution. |
| **Elaboration (EL)** | Builds on, elaborates, evaluates, clarifies own or other's contribution (if own, it should be on separate turns) within an exchange. This adds substantive new information or a new perspective beyond anything said in previous turns, even by one word. |
| **Reasoning Invitation (REI)** | Explicitly invites explanation/justification of a contribution or speculation (new scenarios) /prediction/hypothesis. |
| **Reasoning (RE)** | Provides an explanation or justification of own or another's contribution. Includes drawing on evidence (e.g. identifying language from a text/poem that illustrates something), drawing analogies (and giving reasons for them), making distinctions, breaking down or categorising ideas. |
| **Co-ordination Invitation (CI)** | Invites synthesis, summary, comparison, evaluation or resolution based on two or more contributions (i.e. invites all descriptors of SC and RC below) |
| **Simple Co-ordination (SC)** | Synthesises or summarises collective ideas (at least two, including own and/or others' ideas). Compares or evaluates different opinions, perspectives and beliefs. Proposes a resolution or consensus view after discussion. |
| **Reasoned Co-ordination (RC)** | Compares, evaluates, resolves two or more contributions in a reasoned fashion (e.g. 'I agree with Susan because her idea has more evidence behind it than Emma's'). |
| **Agreement (A)** | Explicit acceptance of or agreement with a statement(s) (e.g. 'Brilliant', 'Good', 'Yeah', 'Okay', I agree with X…). |
| **Querying (Q)** | Doubting, full/partial disagreement, challenging or rejecting a statement. Challenging should be evident through verbal means. |
| **Reference Back (RB)** | Introduces reference to previous knowledge, beliefs, experiences or contributions (includes procedural references) that are common to the current conversation participants. This should refer to a specific activity or time point, not just simple recall (e.g. 'Do you remember what we call it?'). |
| **Reference to Wider Context (RW)** | Making links between what is being learned and a wider context by introducing knowledge, beliefs, experiences or contributions from outside of the subject being taught, classroom or school. |
| **Structural Silence (SU)** | Students may feel 'silenced', such type of silence may be linked to social situations and interpersonal interactions. |
| **Strategic Silence (SA)** | Students choose not to express or articulate an utterance. This experience of strategic silence may be more personal in nature, and although the motivation to self-silence may be influenced by the interactions of others, the decision to remain silent remains at a more private level. |
| **Other Invitation (OI)** | Invitations cannot be coded as any code related to invitation provided above. |
| **Other (O)** | Dialogue turns cannot be coded as any code provided above, |

* This coding scheme is revised from Cambridge Dialogue Analysis Scheme (CDAS, Howe et al., 2019).

## 3.2 Articulating Dialogic Sequences and Rules Development

The primary objective of this stage is to establish an expert rule base for dialogue sequences, which will serve as the foundation for subsequent analyses. This involves synthesising dialogue types and sequences from a comprehensive literature review, classifying them based on their pedagogical goals, and identifying gaps in the existing framework to ensure a robust and inclusive rule base.

***Literature Review and Identification of Core Dialogue Sequences***

To build the expert rule base, we conducted a systematic review of 30 peer-reviewed studies focused on classroom dialogue. These studies span various instructional contexts, covering key pedagogical frameworks such as exploratory talk (Mercer & Wegerif, 1999), dialogic teaching (Alexander, 2008), accountable talk (Michaels et al., 2008), and Socratic dialogue (Dillon, 1988). Each study was analysed to extract its core teaching goals and associated dialogue sequences. Table A1 (see Appendix) summarises the reviewed studies, their core teaching goals, and the dialogue sequences they propose.

***Classification of Dialogue Sequences***

The extracted sequences were then classified into broader categories based on their pedagogical functions. This process involved grouping sequences that shared similar instructional objectives, such as fostering critical thinking, promoting collaboration, or encouraging metacognitive reflection. For instance, sequences like "Reasoning Invitation → Reasoning" and "Elaboration Invitation → Elaboration" were classified under the broader category of *Critical Inquiry*, as they both aim to deepen students' reasoning and analytical skills. Similarly, sequences promoting collaborative construction of knowledge, such as "Simple Co-ordination → Agreement" and "Reasoned Co-ordination → Elaboration," were grouped under *Collaborative Learning*.

This classification process was iterative, involving multiple rounds of discussion and refinement to ensure consistency and coherence. An initial draft of the categorisation was independently reviewed by two domain experts, and discrepancies were resolved through consensus.

***Identification of Gaps in Dialogue Sequences***

During the review, we identified gaps where certain pedagogical goals, such as fostering student agency or addressing socio-emotional aspects of learning, were underrepresented. For example, few studies provided explicit sequences for dialogues aimed at promoting self-regulation or emotional engagement. To address these gaps, we proposed new sequence categories based on theoretical frameworks, such as Zimmerman's model of self-regulated learning (Zimmerman, 2002), and extended existing categories to include socio-emotional components like *Strategic Silence* and *Reference to Wider Context.*

***Validation of the Rule Base Development Process***

To validate the effectiveness of this process, we employed two strategies. First, a panel of five educational researchers reviewed the refined rule base to evaluate its comprehensiveness and theoretical alignment. Their feedback on the clarity and appropriateness of the categories informed further revisions. Second, we conducted an inter-rater reliability test, where two independent coders applied the expert rule base to a subset of classroom dialogue data. The Cohen's Kappa values for each category were above 0.85, indicating high consistency. While this demonstrates the clarity and usability of the rule definitions, it only indirectly supports the validity of the categorisation itself.

This rigorous validation process ensures that the expert rule base not only captures the complexity of classroom dialogue but also aligns closely with established pedagogical theories.

### 3.3 Automated Annotation Agent Development and Validation

To automate the annotation process, we developed an LLM-based agent that recognises sequences in classroom dialogues using the expert rules defined in the rule base. This section outlines the methods used for validating the agent's performance, focusing on inter-coder reliability and time cost efficiency.

***Inter-Coder Reliability Evaluation***

To assess the consistency of the automated tool in applying expert-defined rules, we employed Cohen's Kappa, a widely recognised metric for inter-coder reliability. This metric measures the level of agreement between two coders, with values ranging from 0 to 1. Kappa values above 0.75 indicate strong agreement. In this study, Cohen's Kappa was used to compare the automated tool's categorisation of dialogue sequences with those of human coders across multiple dialogue categories.

***Time Cost and Efficacy Assessment***

We evaluated the efficiency of the automated tool by comparing its coding speed with that of manual annotation. For this purpose, human coders and the automated tool were tasked with coding the same dataset, which comprised 1,084 dialogue turns. The time taken by each method to complete the annotation was recorded to assess the potential time savings offered by the automated tool. This evaluation also considered the tool's scalability and cost-effectiveness for large-scale educational research.

## 4. Results

### 4.1 Consolidating Classroom Dialogue Types Based on Previous Research

To address the first research objective, we synthesised various types of classroom dialogues identified in the 30 reviewed studies to develop a comprehensive framework that captures the diversity and complexity of classroom interactions. This framework is grounded in both the pedagogical goals and the dialogue sequences extracted from the literature. The consolidation process involved categorising these sequences into four primary dialogue types based on their instructional purposes: *Critical Inquiry*, *Collaborative Construction of Knowledge*, *Instructional and Supportive Dialogue*, and *Reflective and Metacognitive Dialogue*. These categories are summarised in Table 2, which lists the corresponding teaching goals and key sequences associated with each type.

**Table 2. Consolidated categories of classroom dialogue based on previous research**

| Consolidated categories | Explanations | Classroom dialogue types in previous research |
|---|---|---|
| Critical Inquiry | Involves dialogue types that focus on critical thinking, questioning, and reasoning.<br>All these dialogue types involve a high level of critical thinking, questioning, and reasoning. They aim to develop deeper understanding and analytical skills through structured questioning and argumentation. | Exploratory Talk<br>Socratic Dialogue<br>Dialectical Dialogue<br>Inquiry-Based Dialogue<br>Accountable Talk |
| Collaborative Construction of Knowledge | Emphasises building knowledge collectively through supportive and collaborative interactions.<br>These dialogue types emphasize building knowledge collectively. Participants build on each other's contributions in a supportive and collaborative environment, enhancing mutual understanding and collective learning. | Cumulative Talk<br>Collaborative Dialogue<br>Interactive Whole-Class Teaching |
| Instructional and Supportive Dialogue | Teacher-led dialogues that provide information, check understanding, and support learning.<br>These dialogue types are teacher-led and focus on providing information, checking understanding, and supporting | Instructional Talk<br>Cognitive Scaffolding<br>Interrogative Dialogue |
| | students' learning processes. They include structured questioning and instructional support to guide students. | |
| Reflective and Metacognitive Dialogue | Encourages deep reflection on learning and teaching processes, fostering metacognitive skills.<br>Both types involve deep reflection on learning experiences and teaching practices. They encourage metacognitive skills by prompting students and teachers to think about their thinking and learning processes. | Reflective Dialogue |
| | | Reflective Teaching |

***Critical Inquiry***

Critical Inquiry involves dialogue types that focus on critical thinking, questioning, and reasoning. These dialogues aim to develop a deeper understanding and analytical skills through structured questioning and argumentation. For example, Exploratory Talk is a type of dialogue where participants engage critically but constructively with each other’s ideas, challenging and counter-challenging contributions in a collaborative spirit (Mercer & Wegerif, 1999). Similarly, Socratic Dialogue stimulates critical thinking and illuminates ideas through asking and answering questions (Dillon, 1988). Dialectical Dialogue involves reasoning through structured argumentation, comparing and contrasting opposing viewpoints to reach a higher level of understanding (Bakhtin, 1981). Inquiry-Based Dialogue centres around asking questions, investigating, and building knowledge through exploration and discovery (Wells, 1999). Lastly, Accountable Talk ensures that dialogue is accountable to the learning community, accurate knowledge, and rigorous thinking, making it purposeful and coherent to support learning (Michaels, O’Connor, & Resnick, 2008).

***Collaborative Construction of Knowledge***

This category emphasises building knowledge collectively through supportive and collaborative interactions. Participants build on each other’s contributions in a supportive environment, enhancing mutual understanding and collective learning. Cumulative Talk is characterised by speakers building positively but uncritically on what others have said, with mutual agreement and repetition (Mercer, 1995). Collaborative Dialogue involves learners working together towards a common goal, sharing knowledge and ideas to solve problems collectively (Rogoff, 1990). Additionally, Interactive Whole-Class Teaching involves the entire class in interactive discussions, where everyone contributes to building knowledge collectively.

### *Instructional and Supportive Dialogue*

This category includes teacher-led dialogues that provide information, check understanding, and support learning processes. These dialogues are characterised by structured questioning and instructional support. Instructional Talk is teacher-directed talk aimed at transmitting information and checking understanding, typically characterised by a sequence of teacher questions and student responses (Cazden, 2001). Cognitive Scaffolding involves the teacher providing support to students as they develop new skills or understanding, gradually withdrawing support as students become more competent (Wood, Bruner, & Ross, 1976). Interrogative Dialogue is a structured form of dialogue characterised by systematic questioning aimed at probing deeper into the subject matter (Hyman, 1979).

### *Reflective and Metacognitive Dialogue*

Reflective and Metacognitive Dialogue encourages deep reflection on learning experiences and teaching practices, fostering metacognitive skills. Reflective Dialogue involves participants thinking deeply about their learning experiences and the concepts being discussed, promoting metacognition and self-assessment (Schön, 1987). Reflective Teaching involves teachers and students engaging in reflective practices to examine and improve teaching and learning processes (Dewey, 1933).

These consolidated categories and their corresponding dialogue types provide a structured framework for analysing and understanding the variety of interactions that occur in educational settings. This framework guides the subsequent analysis and validation of our expert rule-based AI agent. By systematically categorising classroom dialogues into Critical Inquiry, Collaborative Construction of Knowledge, Instructional and Supportive Dialogue, and Reflective and Metacognitive Dialogue, we ensure a comprehensive and robust approach to understanding and enhancing classroom interactions. Detailed explanations of these categories and their dialogue types are provided in the appendix.

## 4.2 Developing rules for the expert rule-based AI agent

To categorise classroom dialogues effectively, we developed an expert rule base that consolidates theoretical and empirical insights into four key dialogue categories: Critical Inquiry, Collaborative Construction of Knowledge, Instructional and Supportive Dialogue, and Reflective and Metacognitive Dialogue. Each category is defined by a meta-rule that guides its identification, key features that characterise its structure, and illustrative examples demonstrating its practical application. These elements ensure the framework's applicability across diverse educational settings.

Table 3. Expert rules for evaluating the consolidated classroom dialogue categories

| Consolidated categories | Meta-rule | Specific rules | Possible sequences | Examples |
|---|---|---|---|---|
| Critical Inquiry | Identify instances where participants challenge and counter-challenge ideas constructively. | If a dialogue has more than three dialogic turns in one topic,<br>and includes structured questioning (REI, ELI) and argumentation (RE, EL),<br>and contains the querying (Q) action,<br>then classify the dialogue as critical inquiry. | REI → RE → Q:<br>Reasoning invitation followed by reasoning and then querying the reasoning.<br>Q → RE → REI:<br>Challenging a statement followed by reasoning and then inviting further reasoning.<br>CI → Q → RE:<br>Co-ordination invitation followed by querying and then reasoning.<br>ELI → Q → RE:<br>Elaboration invitation followed by challenging and then reasoning. | REI→ RE → Q:<br>Teacher: Why do you think…?<br>Student: Because…<br>Teacher: Based on what you say, can you further explain…?<br><br>CI → Q → RE:<br>Teacher: According to their answers, can you explain why…?<br>Student: Because…<br><br>ELI → Q → RE:<br>Teacher: Could you please illustrate your idea and give us a reason?<br>Student: I think…because… |
| Collaborative Construction of Knowledge | These dialogues emphasise mutual understanding and collective learning, where participants contribute constructively to the conversation. | If a teacher-student dialogue involves at least two students discussing the same topic,<br>and the dialogue includes Simple Coordination (SC) or Reasoned Coordination (RC), and Agreement (A), then classify the dialogue as collaborative construction of knowledge.<br>If a student-student dialogue involves at least three students discussing the same topic,<br>and the dialogue includes Simple Coordination (SC) or Reasoned Coordination (RC), and Agreement (A), then classify the dialogue as collaborative construction of knowledge. | ELI → EL → SC/RC:<br>Elaboration invitation followed by elaboration and then simple or reasoned co-ordination.<br>SC/RC → EL → A:<br>Simple or reasoned co-ordination followed by elaboration and then agreement.<br>ELI → A → SC/RC:<br>Elaboration invitation followed by agreement and then simple or reasoned co-ordination.<br>A → EL → SC/RC:<br>Agreement followed by elaboration and then simple or reasoned co-ordination. | ELI → EL → SC/RC:<br>Teacher: Could you expand on your idea further and explain your reasoning?<br>Student: Sure, I think... because...<br>Teacher: Great, now let's see how this explanation ties back to the main concept we discussed.<br>SC/RC → EL → A:<br>Teacher: So, based on this coordination, can you elaborate more on how it works?<br>Student: Yes, it works by… because…<br>Teacher: I agree, your reasoning is sound and connects well with the framework.<br>ELI → A → SC/RC:<br>Teacher: Can you explain your thought process here?<br>Student: Yes, I think I understand the concept and I agree with what we discussed earlier.<br>Teacher: Good, now let's coordinate this agreement with another example to solidify your understanding.<br>A → EL → SC/RC:<br>Teacher: Do you agree with the statement provided?<br>Student: Yes, I agree. I would also add that...<br>Teacher: Excellent! Now let's see how we can link this elaboration with the previous findings. |
| Instructional and Supportive Dialogue | Identify instances of structured questioning and instructional support provided by the teacher. | If a dialogue sequence includes consecutive occurrences of Other Invitation (OI) and Other (O),<br>or an Other Invitation (OI) is followed by a topic switch without a response, then classify the dialogue as Instructional and Supportive Dialogue. | OI → ELI → EL:<br>Instructional talk followed by elaboration invitation and then elaboration.<br>REI → RE → OI:<br>Reasoning invitation followed by reasoning and then instructional talk.<br>ELI → EL → OI:<br>Elaboration invitation followed by elaboration and then instructional talk. | OI → ELI → EL:<br>Teacher: Let's review the steps to solve this equation first. (Instructional talk)<br>Teacher: Can you expand on why we apply this method here? (Elaboration invitation)<br>Student: We use this method because… it simplifies the problem. (Elaboration)<br>REI → RE → OI:<br>Teacher: Why do you think this approach works best? (Reasoning invitation)<br>Student: Because it reduces the complexity of the solution… (Reasoning)<br>Teacher: That's right! Now let me explain how this method fits into the broader concept. (Instructional talk)<br>ELI → EL → OI:<br>Teacher: Can you explain your thought process in more detail? (Elaboration invitation)<br>Student: Sure, I thought this method would work because it aligns with the principles we discussed… (Elaboration)<br>Teacher: Exactly! Let me now clarify how this ties into the next section of our lesson. (Instructional talk) |
| Reflective and Metacognitive Dialogue | Look for instances where participants engage in deep reflection and self-assessment. | If a dialogue is initiated by either a teacher or a student, and includes Refer Back (RB) or Reference to Wider Context (RW),<br>then classify the dialogue as Reflective and Metacognitive Dialogue. | REI → RE → RB/RW:<br>Reasoning invitation followed by reasoning and then refer back.<br>RB → EL → RW:<br>Reflective turn followed by elaboration and then refer to wider context.<br>CI → RB/RW → SC:<br>Co-ordination invitation followed by reflective turn and then simple co-ordination.<br>RB/RW → ELI → EL:<br>Reflective turn followed by elaboration invitation and then elaboration. | REI → RE → RB/RW:<br>Teacher: Why did you choose this particular strategy? (Reasoning invitation)<br>Student: I chose it because it seemed to work well with similar problems. (Reasoning)<br>Teacher: Let’s revisit what we discussed earlier about this strategy’s application in other contexts. (Refer back)<br><br>RB → EL → RW:<br>Teacher: Earlier, you mentioned that this method worked well. How does that align with your current thinking? (Reflective turn)<br>Student: Now that I think about it, it seems to make sense because... (Elaboration)<br>Teacher: This highlights how important it is to consider this approach in a broader range of cases. (Refer to wider context)<br><br>CI → RB/RW → SC:<br>Teacher: Can you connect this solution with another concept we’ve covered? (Co-ordination invitation)<br>Student: When reflecting on both, I see they share a common method of simplifying problems. (Reflective turn) |
| | | | | Teacher: So, both use simplification as a key strategy. (Simple co-ordination)<br><br>RB/RW → ELI → EL:<br>Teacher: Think about how this approach compares to what we studied earlier. (Reflective turn)<br>Teacher: Can you explain in more detail why this comparison is important? (Elaboration invitation)<br>Student: Well, this approach works better because it... (Elaboration) |

As shown in Table 3, critical inquiry dialogues involve deep engagement with ideas through questioning, reasoning, and structured argumentation. The goal of critical inquiry is to develop analytical skills and deeper understanding. Hence, the meta-rule for allocating critical inquiry is to identify instances where participants constructively challenge and counter-challenge ideas. Specifically, the rule for evaluating whether a dialogue qualifies as critical inquiry is:

**If** a dialogue has more than three dialogic turns on one topic, **and** includes structured questioning (REI, ELI) and argumentation (RE, EL), and contains the querying (Q) action, **then** classify the dialogue as critical inquiry.

Therefore, the crucial codes for critical inquiry include Reasoning Invitation (REI), Reasoning (RE), Elaboration Invitation (ELI), Elaboration (EL), and Querying (Q). The conventional critical inquiry dialogue sequences are as follows:

- REI → RE → Q: Reasoning invitation followed by reasoning and then querying the reasoning.
- Q → RE → REI: Challenging a statement followed by reasoning and then inviting further reasoning.
- CI → Q → RE: Co-ordination invitation followed by querying and then reasoning.
- ELI → Q → RE: Elaboration invitation followed by challenging and then reasoning.

Collaborative construction of knowledge dialogues emphasis mutual understanding and collective learning, where participants contribute constructively to the conversation. The goal is to enhance mutual understanding and collective learning by allowing participants to build on each other's contributions positively. Hence, the meta-rule for allocating collaborative construction of knowledge is to identify instances where participants constructively build on and support each other's ideas. Specifically, the rule for evaluating whether a dialogue qualifies as collaborative construction of knowledge is:

**If** a teacher-student dialogue involves at least two students discussing the same topic, **and** the dialogue includes Simple Coordination (SC) or Reasoned Coordination (RC), and Agreement (A), **then** classify the dialogue as collaborative construction of knowledge.

**If** a student-student dialogue involves at least three students discussing the same topic, **and** the dialogue includes Simple Coordination (SC) or Reasoned Coordination (RC), and Agreement (A), **then** classify the dialogue as collaborative construction of knowledge.

Therefore, the crucial codes for this category include Simple Coordination (SC), Reasoned Coordination (RC), and Agreement (A). The conventional collaborative construction of knowledge dialogue sequences are as follows:

- ELI → EL → SC/RC: Elaboration invitation followed by elaboration and then simple or reasoned co-ordination.
- SC/RC → EL → A: Simple or reasoned co-ordination followed by elaboration and then agreement.
- ELI → A → SC/RC: Elaboration invitation followed by agreement and then simple or reasoned co-ordination.
- A → EL → SC/RC: Agreement followed by elaboration and then simple or reasoned co-ordination.

Instructional and supportive dialogues are characterised by teacher-led interactions that provide information, check understanding, and support learning processes. The goal of these dialogues is to ensure student comprehension and facilitate learning through structured questioning and instructional support. Hence, the meta-rule for allocating instructional and supportive dialogue is to identify instances where the teacher provides clear instructional guidance and support. Specifically, the rule for evaluating whether a dialogue qualifies as instructional and supportive dialogue is:

**If** a dialogue sequence includes consecutive occurrences of Other Invitation (OI) and Other (O), **or** an Other Invitation (OI) is followed by a topic switch without a response, **then** classify the dialogue as Instructional and Supportive Dialogue.

Therefore, the crucial codes for this category include Other Invitation (OI) and Other (O). The conventional instructional and supportive dialogue sequences are as follows:

- OI → ELI → EL: Instructional talk followed by elaboration invitation and then elaboration.
- REI → RE → OI: Reasoning invitation followed by reasoning and then instructional talk.
- ELI → EL → OI: Elaboration invitation followed by elaboration and then instructional talk.

Reflective and metacognitive dialogues encourage deep reflection on learning experiences and promote metacognitive skills. The goal of these dialogues is to enhance understanding and self-regulation by prompting participants to think about their thinking and learning processes. Hence, the meta-rule for allocating reflective and metacognitive dialogue is to identify instances where participants engage in deep reflection and self-assessment. Specifically, the rule for evaluating whether a dialogue qualifies as reflective and metacognitive dialogue is:

**If** a dialogue is initiated by either a teacher or a student, and includes Refer Back (RB) or Reference to Wider Context (RW), **then** classify the dialogue as Reflective and Metacognitive Dialogue.

Therefore, the crucial codes for this category include Refer Back (RB) and Reference to Wider Context (RW). The conventional reflective and metacognitive dialogue sequences are as follows:

- REI → RE → RB/RW: Reasoning invitation followed by reasoning and then refer back.
- RB → EL → RW: Reflective turn followed by elaboration and then refer to wider context.
- CI → RB/RW → SC: Co-ordination invitation followed by reflective turn and then simple co-ordination.
- RB/RW → ELI → EL: Reflective turn followed by elaboration invitation and then elaboration.

### 4.3 Developing Expert Rule-Based AI Agent for Sequence Analysis and Validation

#### The developing process of the expert rule-based AI agent

The expert rule-based AI agent for sequence analysis was developed by building upon the automatic dialogue analysis agent proposed by authors (2024). Expert rules were systematically fed into the previous agent, enabling it to recognise specific dialogue sequences across various categories. The development process focused on ensuring that the agent accurately applied these rules to categorise classroom dialogues effectively.

#### Data collection and evaluation framework

The sample for this study consisted of classroom dialogue data from a middle school in China, comprising 12 classes, including both Chinese and mathematics lessons. A total of 1,084 dialogue turns were recorded and analysed. To validate the performance of the expert rule-based AI agent, human coders manually annotated the same dataset for comparison.

The evaluation framework centred on several key metrics: accuracy, consistency, time cost, scalability, and cost-effectiveness. Accuracy and consistency were measured using precision with comparisons made between the agent's outputs and human coders' annotations. Cohen's Kappa was employed to assess inter-coder reliability. Performance in terms of speed and consistency was evaluated by comparing the agent's processing time with that of human coders, while scalability and efficacy were gauged through the agent's ability to handle large datasets efficiently. Finally, cost-effectiveness was assessed by comparing the reduction in labour hours and associated expenses with traditional manual coding methods.

**Accuracy and consistency**

Precision Comparison with Human Coders

The coding performance of the coding tool was compared with human coders using common evaluation metrics, mainly including precision, to assess its accuracy in applying the coding scheme to classroom dialogues.

Precision measures the proportion of the tool correctly identifies and categorises dialogue sequences that match human annotations. In this case, the tool demonstrated high precision across several categories. For instance, the tool achieved a precision of 97.3% for coding critical inquiry dialogues, meaning that when the tool coded a dialogue as constructively challenging interlocutors, it matched human coders 97.3% of cases. Additionally, the tool achieved a precision of 95.6% for coding collaborative construction of knowledge dialogues. This indicates that when the tool identified a dialogue as promoting shared understanding or knowledge co-construction, it agreed with human coders 95.6% of the time. For instructional and supportive dialogue, the tool demonstrated a high precision of 98.1%, meaning that it correctly matched human annotations 98.1% of the time when identifying dialogues that provided instruction or support. Lastly, the tool achieved a precision of 96.5% for coding reflective and metacognitive dialogues, showing a 96.5% agreement with human coders when categorising dialogues focused on reflection or metacognition.

Table 4. Precision Rates for Each Dialogue Categories

| Dialogue Categories | Precisions | Cohen's Kappa |
|---|---|---|
| Critical Inquiry | 97.3% | 0.952*** |
| Collaborative Construction of Knowledge | 95.6% | 0.871*** |
| Instructional and Supportive Dialogue | 98.1% | 0.914*** |
| Reflective and Metacognitive Dialogue | 96.5% | 0.895*** |

Inter-Coder Reliability

The automated agent demonstrated substantial agreement with human coders across various dialogue categories, achieving a Cohen's Kappa score of 0.952 in categorising critical inquiry dialogues. This exceptionally high score reflects the agent's strong ability to reliably classify complex dialogue interactions, closely matching human annotations. Additionally, the tool achieved a Cohen's Kappa of 0.871 for collaborative construction of knowledge dialogues, indicating robust agreement with human coders in identifying shared understanding or co-construction of knowledge. For instructional and supportive dialogue, the agent reached a Kappa score of 0.914, further demonstrating its consistency in coding dialogues that offer instructional guidance or support. Lastly, the tool achieved a Cohen's Kappa of 0.895 for reflective and metacognitive dialogues, showing that it closely aligned with human annotations in recognising reflective and metacognitive interactions.

The high Kappa scores across multiple dialogue types, including elaboration invitations and reasoning sequences, underscore the robustness of the expert rule base. These results indicate that the tool consistently applies the coding framework in a wide range of classroom contexts, reducing variability and ensuring reliable performance regardless of the complexity of the dialogues being analysed.

Performance in Speed and Consistency

The automated tool consistently outperformed manual coding in both speed and consistency, particularly in handling complex dialogue sequences. For example, in coding intricate patterns like elaboration followed by reasoning (dialogues include ELI → EL → RE sequence), the tool applied the coding rules with an accuracy of 95%, whereas human coders often showed variability, especially in borderline cases where distinctions between elaboration and reasoning were subtle.

Additionally, the tool reduced inconsistencies that often arise from human subjectivity, particularly in dialogue types that require nuanced interpretations, such as reflective dialogue and instructional talk. Human coders may differ in their interpretation of these subtle distinctions, but the automated agent applied the coding rules consistently, ensuring that the same criteria were used across all dialogues. This consistency enhances the reliability of the analysis, leading to more robust and reproducible research outcomes.

**Time cost and efficacy**

Comparison of Time Taken

The automated tool provides significant time savings compared to manual coding. On average, manually coding a classroom session with 100 dialogue turns took a human coder approximately 3.5 to 4 hours, while the automated tool processed the same data in just 8 to 10 minutes, representing a 96% reduction in coding time. This drastic improvement in efficiency allows for faster data processing and analysis, freeing up valuable research time and resources.

In a study involving the analysis of 12 classroom sessions with a total of 1,089 dialogue turns, manual coding required around 42 hours to complete. In contrast, the automated tool finished the same task in approximately 130 minutes (just over 2 hours), demonstrating a 94.8% reduction in coding time. These time savings make the automated tool an invaluable resource for large-scale educational research, enabling more extensive studies to be completed within a shorter timeframe.

Scalability and Efficacy

The introduction of automation in dialogue coding significantly enhances the scalability of research, allowing researchers to handle much larger datasets than was feasible with manual methods. With the automation of coding, researchers can now analyse dialogue data from hundreds of classrooms in a fraction of the time, making large-scale, longitudinal studies both practical and achievable. This scalability enables the collection and analysis of data from diverse educational settings, providing a more comprehensive understanding of classroom interactions.

In addition, the tool's ability to process large datasets improves the generalizability of research findings. Traditional manual coding often limits studies to smaller sample sizes due to the significant time and effort required. With automation, however, researchers are no longer constrained by these limitations and can analyse data from large, varied samples, resulting in findings that are more widely applicable across different contexts. This scalability not only enhances the robustness of research but also provides deeper insights into classroom dialogue patterns, leading to more reliable and actionable conclusions for educators and policymakers.

## 5. Conclusion & Discussion

### 5.1 Summary of results

This study achieved two significant outcomes. First, we developed a comprehensive expert knowledge base specifically designed for classroom dialogue sequence analysis. To the best of our knowledge, this knowledge base synthesises and categorises a large body of research, drawing from more than 30 published studies. It provides an extensive classification of dialogue categories, offering a detailed and exhaustive analysis of the various types of interactions observed in classroom settings. The thoroughness of this classification ensures that the knowledge base can accommodate a broad range of dialogue sequences, making it an invaluable resource for future studies on classroom discourse.

Second, we successfully developed an LLM-based agent that integrates the advantages of rule-based systems and general AI. By combining precise, expert-informed rules with the adaptability and pattern-recognition capabilities of general AI, the agent is capable of accurately identifying and categorising dialogue sequences while maintaining flexibility to adapt to the nuances of natural language. This dual capability addresses a critical research gap identified in the literature, where previous methods either lacked theoretical grounding or failed to effectively incorporate expert knowledge into computational models. Our agent bridges this gap by providing a theoretically informed, yet adaptive, solution for automating classroom dialogue analysis. This not only ensures alignment with pedagogical principles but also enhances the model's robustness in handling the complexity and variability of real-world classroom discourse, advancing both theory-driven and data-driven approaches.

### 5.2 Signification of the study

The significance of this research is twofold. First, this study represents a significant advancement in applying rule-based AI systems to the analysis of classroom dialogue sequences. While rule-based AI has been applied in other domains, its introduction into classroom interaction research is relatively novel. The combination of an expert-informed rule-based system with advanced LLMs enables a more structured and theoretically grounded analysis of dialogue patterns. This approach can deepen our understanding of classroom dynamics by ensuring that the agent's categorisations are aligned with established educational theories, offering more accurate and reliable insights compared to methods driven purely by data patterns.

Second, although the current expert knowledge base is comprehensive, it remains adaptable and open to future refinement. The structure of this knowledge base allows for continual improvement as more data becomes available and new insights emerge. Researchers have the opportunity to collaboratively expand the knowledge base, incorporating a greater variety of dialogue sequences and creating more nuanced subcategories within existing categories. This ongoing expansion enhances the diversity of the sequences covered and improves the robustness of the dialogue categorisation process. In this way, the expert knowledge base can evolve to meet the growing complexity of classroom dialogue analysis.

In summary, in terms of theoretical value, this research introduces a new methodology for analysing classroom dialogue, providing a scalable and systematic approach that is deeply rooted in educational theory. Practically, the development of this agent has significant implications for educators, as it enables real-time feedback on classroom interactions. By offering immediate insights into the quality of dialogue, the agent allows teachers to adjust their instructional strategies dynamically, thereby improving student outcomes. Additionally, the foundation laid by this study opens the door for future research aimed at refining and expanding the expert rule base, further enhancing the accuracy and applicability of classroom dialogue analysis in various educational contexts.

### 5.3 Future Directions

Looking towards the future, several avenues for further research and development emerge from this study. One promising direction is the continual expansion and refinement of the expert knowledge base. As more data from diverse classroom settings becomes available, researchers can collaboratively work to refine the categories and subcategories within the knowledge base. This will allow for a more nuanced and comprehensive understanding of dialogue sequences, especially as educational contexts continue to evolve and diversify. By building a richer knowledge base, future research can address increasingly complex classroom interactions, accommodating a wider range of teaching styles and learning environments.

The agent's ability to categorise and analyse large volumes of classroom dialogue opens the possibility of aggregating these results into a comprehensive quantitative database. This approach enables the integration of qualitative insights with quantitative methods, allowing for large-sample statistical analyses that can uncover patterns and correlations across diverse educational settings. By bridging qualitative depth with quantitative breadth, such a database could support meta-analyses, trend identification, and the validation of pedagogical theories on an unprecedented scale. These findings would provide actionable insights for both researchers and educators, enhancing evidence-based teaching practices. Furthermore, as the agent evolves, it could be integrated into teacher training programs to offer low-stakes, formative feedback for early-career educators. This feedback would help teachers refine their instructional strategies and foster deeper student engagement, leveraging real-time insights without the pressure of high-stakes evaluations. Together, these capabilities position the agent as a powerful tool for advancing both research and practice in education.

In conclusion, the expert rule-based agent and knowledge base developed in this study lay the groundwork for future theoretical advancements and practical applications in classroom dialogue analysis. By continuing to refine and expand the agent, researchers and educators alike can benefit from a more sophisticated understanding of classroom interactions, ultimately improving teaching quality and student learning experiences.

## CRediT authorship contribution statement

**Yun Long**: Writing – original draft, Methodology, Formal analysis, Conceptualization. **Yu Zhang**: Writing – review & editing, Methodology, Conceptualization, Supervision.

## Data availability

Data will be made available on request.

**Appendix**

Table A1. Summary of types of teacher-student dialogue based on previous literature

| Types of teacher-student dialogue | Definition | Reference |
|---|---|---|
| Exploratory Talk | A type of dialogue where participants engage critically but constructively with each other's ideas. Contributions are challenged and counter-challenged but in a collaborative spirit. | Mercer, N., & Wegerif, R. (1999). Is 'exploratory talk' productive talk? In K. Littleton & P. Light (Eds.), *Learning with computers: Analyzing productive interaction* (pp. 79-101). Routledge. |
| Cumulative Talk | A type of dialogue where speakers build positively but uncritically on what each other has said. There is a mutual agreement and repetition without much questioning or critical evaluation. | Mercer, N. (1995). *The guided construction of knowledge: Talk amongst teachers and learners*. Multilingual Matters. |
| Disputational Talk | Characterized by disagreement and individualized decision-making. Participants are more competitive than cooperative, and there is little constructive engagement with others' ideas. | Mercer, N. (1995). *The guided construction of knowledge: Talk amongst teachers and learners*. Multilingual Matters. |
| Dialogic Teaching | Emphasizes the role of dialogue in teaching, where students and teachers engage in back-and-forth communication to co-construct knowledge. | Nystrand, M., Gamoran, A., Kachur, R., & Prendergast, C. (1997). *Opening dialogue: Understanding the dynamics of language and learning in the English classroom*. Teachers College Press. |
| Dialogic Teaching | Emphasizes dialogue through which teachers and students think and reason together. It involves open questions, extended exchanges, and the exploration of ideas. | Alexander, R. J. (2008). *Towards dialogic teaching: Rethinking classroom talk* (4th ed.). Dialogos. |
| Instructional Talk | Teacher-directed talk aimed at transmitting information and checking understanding. It is usually characterized by teacher questions and student responses (IRE/F: Initiate-Response-Evaluate/Feedback). | Cazden, C. B. (2001). *Classroom discourse: The language of teaching and learning* (2nd ed.). Heinemann. |
| Socratic Dialogue | A form of cooperative argumentative dialogue that stimulates critical thinking and illuminates ideas through asking and answering questions. | Dillon, J. T. (1988). Questioning and teaching: A manual of practice. Teachers College Press. |
| Reflective Dialogue | Involves participants thinking deeply about their learning experiences and the concepts being discussed. This type of dialogue encourages metacognition and self-assessment. | Schön, D. A. (1987). *Educating the reflective practitioner: Toward a new design for teaching and learning in the professions*. Jossey-Bass. |
| Collaborative | Dialogues where learners | Rogoff, B. (1990). |
| Dialogue | work together towards a common goal, sharing knowledge and ideas to solve problems collectively. | *Apprenticeship in thinking: Cognitive development in social context*. Oxford University Press. |
| Cognitive Scaffolding | Dialogue where the teacher provides support to students as they develop new skills or understanding, gradually withdrawing support as students become more competent. | Wood, D., Bruner, J. S., & Ross, G. (1976). The role of tutoring in problem-solving. *Journal of Child Psychology and Psychiatry*, 17(2), 89-100. |
| **Interrogative Dialogue** | A structured form of dialogue characterized by systematic questioning aimed at probing deeper into the subject matter. | Hyman, R. T. (1979). *Strategic questioning*. Prentice Hall. |
| Dialectical Dialogue | Involves reasoning through structured argumentation where opposing viewpoints are compared and contrasted to arrive at a higher level of understanding. | Bakhtin, M. M. (1981). *The dialogic imagination: Four essays* (M. Holquist, Ed.; C. Emerson & M. Holquist, Trans.). University of Texas Press. |
| **Inquiry-Based Dialogue** | Centered around asking questions, investigating, and building knowledge through exploration and discovery. | Wells, G. (1999). *Dialogic inquiry: Towards a sociocultural practice and theory of education*. Cambridge University Press. |
| Scaffolded Interaction | Interactions where teachers provide temporary support structures to assist students in achieving a higher level of understanding or skill than they would achieve independently. | Vygotsky, L. S. (1978). *Mind in society: The development of higher psychological processes*. Harvard University Press. |
| Guided Inquiry | The teacher guides students through the inquiry process by posing questions, providing resources, and scaffolding their exploration and understanding. | Kuhlthau, C. C., Maniotes, L. K., & Caspari, A. K. (2015). *Guided inquiry: Learning in the 21st century*. Libraries Unlimited. |
| Reciprocal Teaching | An instructional activity in which students become the teacher in small group reading sessions. Teachers model, | Palincsar, A. S., & Brown, A. L. (1984). Reciprocal teaching of comprehension-fostering and |
| | then help students learn to guide group discussions using four strategies: summarizing, question generating, clarifying, and predicting. | comprehension-monitoring activities. *Cognition and Instruction, 1*(2), 117-175. |
| Accountable Talk | Dialogue that is accountable to the learning community, to accurate knowledge, and to rigorous thinking, ensuring that talk is purposeful, coherent, and supports learning. | Michaels, S., O'Connor, C., & Resnick, L. B. (2008). Deliberative discourse idealized and realized: Accountable talk in the classroom and in civic life. *Studies in Philosophy and Education, 27*(4), 283-297. |
| Reflective Teaching | Involves teachers and students engaging in reflective practices to examine and improve teaching and learning processes. | Dewey, J. (1933). *How we think: A restatement of the relation of reflective thinking to the educative process*. D.C. Heath and Company. |
| Dialogic Inquiry | Focuses on the process of learning through dialogue, where students and teachers collaboratively construct knowledge through discussion and inquiry. | Wells, G. (1999). *Dialogic inquiry: Towards a sociocultural practice and theory of education*. Cambridge University Press. |
| Collaborative Reasoning | A discussion activity where students engage in reasoned argumentation to explore big questions and complex problems. | Clark, H. H., & Schaefer, E. F. (1989). Collaborating on contributions to conversations. *Language and Cognitive Processes, 4*(1), 19-41. |
| Co-Constructive Dialogue | Dialogues where teachers and students jointly construct understanding through active engagement and negotiation of meaning. | Rogoff, B. (1994). Developing understanding of the idea of communities of learners. *Mind, Culture, and Activity, 1*(4), 209-229. |
| Dialogical Argumentation | Dialogues that involve structured argumentation where students present and critically examine different viewpoints to reach a reasoned conclusion. | Kuhn, D. (1991). *The skills of argument*. Cambridge University Press. |